\documentclass[runningheads]{llncs}
\usepackage[american]{babel}
\usepackage{graphicx}

\usepackage{tikz}
\usepackage{comment}
\usepackage{amsmath,amssymb} 
\usepackage{color}
\usepackage{multirow}
\usepackage{bigstrut}
\usepackage[caption=false]{subfig}
\usepackage{wrapfig}
\usepackage{bm}
\usepackage{multirow}
\usepackage{diagbox}
\usepackage{algorithm}
\usepackage{algorithmic}
\usepackage{mathrsfs}
\usepackage{makecell}
\usepackage[utf8]{inputenc} 
\usepackage[T1]{fontenc}    
\usepackage{hyperref}       
\usepackage{url}            
\usepackage{booktabs}       
\usepackage{amsfonts}       
\usepackage{nicefrac}       
\usepackage{microtype}      
\usepackage{xcolor}         
\usepackage[american]{babel}
\usepackage{CJKutf8}
\newcommand{\ie}{{\emph{i.e.}}}
\newcommand{\eg}{{\emph{e.g.}}}

\usepackage[accsupp]{axessibility}  
\usepackage{footnote}
\usepackage{booktabs}
\usepackage{threeparttable}
\usepackage{orcidlink}

\begin{document}
\pagestyle{headings}
\mainmatter
\def\ECCVSubNumber{5964}  

\title{SEMICON: A Learning-to-hash Solution for Large-scale Fine-grained Image Retrieval} 


\titlerunning{SEMICON}
%
\author{Yang Shen\inst{1,2}\orcidlink{0000-0002-6344-9951} \and
Xuhao Sun\inst{1}\orcidlink{0000-0002-6642-6966} \and
Xiu-Shen Wei\inst{1,2,3}\thanks{Corresponding author. Y. Shen, X. Sun, X.-S. Wei and Jian Yang are also with Key Lab of Intelligent Perception and Systems for High-Dimensional Information of Ministry of Education, and Jiangsu Key Lab of Image and Video Understanding for Social Security, Nanjing University of Science and Technology, China. This work is supported by National Key R\&D Program of China (2021YFA1001100), Natural Science Foundation of Jiangsu Province of China under Grant (BK20210340), the Fundamental Research Funds for the Central Universities (No. 30920041111, No. NJ2022028), CAAI-Huawei MindSpore Open Fund, Beijing Academy of Artificial Intelligence (BAAI), and Postgraduate Research \& Practice Innovation Program of Jiangsu Province (KYCX22\_0463).}\orcidlink{0000-0002-8200-1845} \and
Qing-Yuan Jiang\orcidlink{0000 0002 9214 7960} \and
Jian Yang\inst{1}}
\authorrunning{Y. Shen et al.}
%
\institute{School of Computer Science and Engineering, Nanjing University of Science and Technology, China\\ \and
State Key Laboratory of Integrated Services Networks, Xidian University, China\\ \and
State Key Laboratory for Novel Software Technology, Nanjing University, China\\
\email{\{shenyang\_98,sunxh,weixs,csjyang\}@njust.edu.cn, qyjiang24@gmail.com}
}
\maketitle

\begin{abstract}
In this paper, we propose \underline{\textbf{S}}uppression-\underline{\textbf{E}}nhancing \underline{\textbf{M}}ask based attention and \underline{\textbf{I}}nteractive \underline{\textbf{C}}hannel transformati\underline{\textbf{ON}}~(SEMICON) to learn binary hash codes for dealing with large-scale fine-grained image retrieval tasks. In SEMICON, we first develop a suppression-enhancing mask (SEM) based attention to dynamically localize discriminative image regions. More importantly, different from existing attention mechanism simply erasing previous discriminative regions, our SEM is developed to restrain such regions and then discover other complementary regions by considering the relation between activated regions in a stage-by-stage fashion. In each stage, the interactive channel transformation (ICON) module is afterwards designed to exploit correlations across channels of attended activation tensors. Since channels could generally correspond to the parts of fine-grained objects, the part correlation can be also modeled accordingly, which further improves fine-grained retrieval accuracy. Moreover, to be computational economy, ICON is realized by an efficient two-step process. Finally, the hash learning of our SEMICON consists of both global- and local-level branches for better representing fine-grained objects and then generating binary hash codes explicitly corresponding to multiple levels. Experiments on five benchmark fine-grained datasets show our superiority over competing methods. Codes are available at \url{https://github.com/NJUST-VIPGroup/SEMICON}.

\end{abstract}

\section{Introduction}

The explosive growth of images on the web makes learning-to-hash methods become a promising solution for large-scale image retrieval tasks~\cite{wang2017survey}. The objective of image-based hash learning aims to represent the content of an image by generating a binary code for both efficient storage and accurate retrieval~\cite{hoe2021one}. Most existing deep hashing methods~\cite{li2015feature,hoe2021one,cao2017hashnet,qingyuanAAAI18} merely support image retrieval for generic concepts, \eg, cars or planes, which might fall short of practical demand with the rapidly growing amount of real applications associated with \emph{fine-grained} image retrieval~\cite{vegfru,van2018inaturalist,liu2016deepfashion,airplanes}. Thus, recent works on deep hashing~\cite{exchnet,wei2021fine,ma2020correlation,jin2020deep} have begun to focus on fine-grained retrieval which is required to retrieve images accurately belonging to subordinate categories of a meta-category, \eg, different species of animals or plants~\cite{van2018inaturalist}, rather than a generic (coarse-grained) category.

In the literature, existing generic hashing methods always utilized the outputs of the last CNN feature layer to generate binary hash codes~\cite{hoe2021one,cao2017hashnet}. Then, these generated hash codes naturally correspond to the holistic representations of the retrieved visual objects. On the other side, recent fine-grained hashing methods, some of which had achieved good retrieval accuracy, were proposed to be equipped with additional modules for locating fine-grained objects' parts (\eg, birds tails or dogs heads) by region localization~\cite{ma2020correlation,jin2020deep} or local feature alignment~\cite{exchnet}. It is important to know that these located object parts are crucial for fine-grained vision tasks~\cite{krause2014learning,vedaldi2014understanding}. Eventually, similar to generic hashing methods, existing fine-grained hashing still fuses object- and part-level features as a unified feature, and then generates hash codes based on such unified features.

\begin{figure}[t]
\centering
{\includegraphics[width=1.0\textwidth]{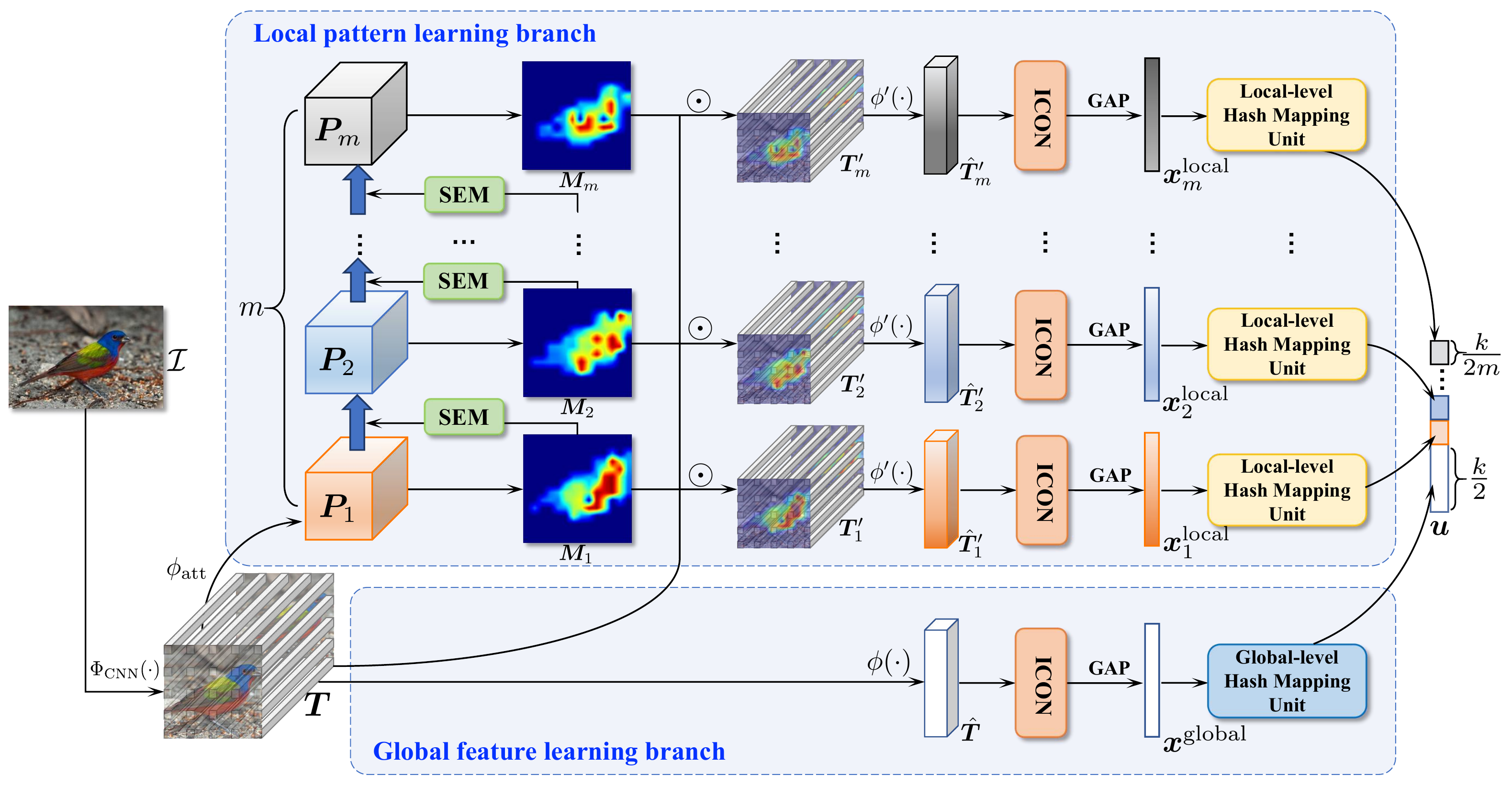}}
\caption{Overall framework of the proposed SEMICON, which consists of two branches, \ie, the global feature learning branch and the local pattern learning branch. In SEMICON, the SEM module is designed to generate $m$ attention maps (\ie, $\bm{M}_i$) stage-by-stage and the ICON module takes each channel as token embeddings to make interactions among different channels. The whole network is end-to-end trainable. 
}
\label{fig:fw}
\end{figure}

Therefore we ask: \emph{What is the explicit meaning of these hash codes}? In order to make the learnt hash codes explicitly meaningful and interpretable, we propose \underline{\textbf{S}}uppression-\underline{\textbf{E}}nhancing \underline{\textbf{M}}ask based attention and \underline{\textbf{I}}nteractive \underline{\textbf{C}}hannel transformati\underline{\textbf{ON}}~(SEMICON), cf. Figure~\ref{fig:fw}. Our SEMICON is designed by having two branches: The one is a global feature learning branch with a single global hashing unit for representing the object-level meanings, while the other one is a local pattern learning branch with multiple local hashing units for representing the multiple (different) part-level meanings in a stage-by-stage fashion. As presented in Figure~\ref{fig:fw}, our final generated hash bits consists of a single object-level hash code and multiple part-level hash codes. Each hash code could explicitly correspond to its own semantic meaning.

In SEMICON, it has two crucial modules, including the suppression-enhancing mask based attention~(SEM) module and the interactive channel transformation~(ICON) module. More specifically, SEM is applied in each learning stage of the local pattern learning branch for dynamically localizing discriminative image regions one-by-one. However, different from other attention-based methods, our SEM is developed to restrain such regions and then discover other complementary regions by considering the relation between activated regions. Therefore, the image regions located in two adjacent stages will be correlated, which will be beneficial to the fine-grained tailored representations. For ICON, this module is employed upon each feature tensor (\eg, $\bm{\hat{T}}$ in Figure~\ref{fig:fw}) by adopting its channels as token embeddings to make interactions across different channels, cf. Section~\ref{sec:ICON}. Since channels can generally correspond to visual object parts~\cite{chen2017sca,liu2015treasure,simon2015neural}, ICON can also model the part correlation accordingly. It could further improve fine-grained retrieval accuracy by considering the internal semantic interactions/correlations among discriminative parts~\cite{ma2020correlation,cai2017higher}. However, as directly calculating the correlations across all channels is computationally complex, we implement this module as a two-step process in order to be efficient and scalable. Extensive experimental results on five benchmark fine-grained retrieval datasets suggest that our method achieves the new state-of-the-art performance.

The main contributions of our work are three-fold. (1) We propose a novel method, \ie, the suppression-enhancing mask based attention and interactive channel transformation, for dealing with the fine-grained hash learning task. (2) We design a suppression-enhancing mask based attention operation to maintain relations between different activated regions and propose a two-step interactive channel transformation module to build correlations between different channels. (3) Experimental results on five benchmark datasets show that our SEMICON achieves significant improvements over competing methods.

\section{Related Work}
\subsection{Fine-Grained Image Retrieval}
Fine-grained retrieval is a fundamental topic of fine-grained image analysis~\cite{wei2021fine} which has gained more and more traction in recent years~\cite{jigsawCVPR20,exchnet,xiawuaaai19,wei20212,song2017deep}. Compared with generic image retrieval, which focuses on retrieving similar images based on similarities in their content~(\eg, texture, color, and shape), fine-grained retrieval aims to retrieve the images of the same category type~(\eg, the same subordinate species of animals~\cite{van2018inaturalist}) with only subtle differences~(\eg, different beak colors or claw shapes of birds). 

Depending on the types of query images, fine-grained image retrieval tasks can be separated into two groups, \ie, fine-grained content based image retrieval (FG-CBIR) and fine-grained sketch-based image retrieval (FG-SBIR). In concretely, SCDA~\cite{wei2017selective} is one of the earliest work of FG-CBIR that used deep pre-trained networks without using explicit localization supervisions. Supervised metric learning based approaches were then proposed to overcome the retrieval accuracy limitations of unsupervised retrieval~\cite{cai2017higher}. In the other research line, FG-SBIR is another interesting task related to both fine-grained image retrieval and cross-modal retrieval of which goal is to match specific photo instances using a free-hand sketch as the query modality. Existing FG-SBIR approaches generally aim to train embedding space where sketches and photos can be compared in a nearest neighbor fashion~\cite{yu2016sketch,song2017deep}.

As all these fine-grained retrieval methods utilize the outputs of the last feature layer of deep networks to deal with retrieval tasks, they still have limitations in the face of large-scale data even if they have achieved good results. To be specific, the searching time for exact nearest neighbor is typically expensive or even impossible for the given queries in large-scale retrieval tasks. To alleviate this issue, fine-grained hashing, which aims to generate compact binary codes to represent fine-grained objects, as a promising direction has attracted the attention in the fine-grained community very recently~\cite{exchnet,jin2020deep,zeng2021pyramid,wei20212}.

\subsection{Learning to Hash}
Hashing has been widely-studied to transform the data item to a short code consisting of a sequence of bits (\ie, hash codes). Compared to data-independent hashing~\cite{fastkdd17th,vldb2007,densifyingicml14}, data-dependent hashing~(\emph{aka} learning to hash) aims to learn hash codes that are more compact yet more data-specific. Due to the discrete of hash codes and non-differentiability of binary hash functions, the optimization of learning to hash is NP-hard~\cite{hoe2021one}. 

Specifically, data-independent hashing methods attempted to adjust hash generating from different perspectives, \eg, the theory or machine learning views, to name a few: proposing random hash functions satisfying local sensitive property~\cite{fastkdd17th}, developing better search schemes~\cite{vldb2007}, providing faster computation of hash functions~\cite{densifyingicml14}, etc. In contrast with data-independent hashing methods, since data-dependent hashing methods learn hash functions from a specific dataset to achieve similarity preserving, they can generally obtain superior retrieval accuracy. Especially for capitalizing on advances in deep learning, many well-performing methods were proposed to integrate feature learning and hash code learning into an end-to-end framework based on deep networks, \eg,~\cite{qingyuanAAAI18,cakir2018hashing,yuan2018relaxation}. 

Fine-grained hashing, as a more challenging and practical hashing task in the vision community, has achieved great attention in very recent years~\cite{exchnet,jin2020deep,wei20212,zeng2021pyramid,lu2021swinfghash}. In the literature, ExchNet~\cite{exchnet} and DSaH~\cite{jin2020deep} defined the fine-grained hashing task almost at the same time. While, they added additional modules to extract local-level features for representing objects' parts, and then aggregated both global-level features and local-level features together to generate the unified binary hash codes. SwinFGHash~\cite{lu2021swinfghash} did not add extra modules but took transformer-based architecture to model the feature interactions. The learnt hash bits of these methods seem incomprehensible and lack semantics which are meaningful to fine-grained objects as we do not know what these hash bits explicitly indicate. Although \textsc{A$^2$-Net}~\cite{wei20212} tried to equip those learnt unified hash codes with correspondence to object attributes~\cite{learnVAnips}, the hash mapping component still mixes up multiple levels of features together which made the hash codes ambiguous w.r.t. clear visual semantics. In this paper, we do not aggregate all the features from different levels together to generate the unified hash codes, but generate the final hash codes corresponding to the features from different levels in a stage-by-stage fashion. 

\subsection{Attention Mechanism}
Attention mechanisms are those methods for diverting attention to the most important regions of an image and disregarding irrelevant regions~\cite{guo2021attention}. In the past years, attention mechanism has played an increasingly important role and has provided benefits in many vision tasks, \eg, image classification~\cite{woo2018cbam}, image retrieval~\cite{ng2020solar} and object detection~\cite{carion2020end}. In a vision system and Deep Neural Networks~(DNNs), an attention mechanism can be viewed as a step of dynamically selecting and adaptively weighting features according to the importance of inputs.

Attention mechanisms can be categorised according to data domain~\cite{guo2021attention}. Besides temporal attention~\cite{li2019global} and branch attention~\cite{li2019selective}, most of the existing attention mechanisms are related to channel information. In DNNs, different channels in different feature maps usually represent different objects' parts~\cite{chen2017sca,liu2015treasure,simon2015neural}. Channel attention adaptively recalibrates the weight of each channel in DNNs which can be viewed as an object selection process~\cite{guo2021attention}.  In fine-grained tasks, researchers often adopt the erasing operation~\cite{zhang2018adversarial,liu2019bidirectional} on the most discriminative regions, which can also be described as the most activated channels, to mine discerning information from the rest of the channels. However, these erasing based attention methods seem less informative that the relations across different regions are completely lost. 

Recently, self-attention, which has achieved great success in Natural Language Processing~\cite{vaswani2017attention}, has also shown the potential to become a dominant tool in vision tasks~\cite{dosovitskiy2020image,liu2021swin}. Typically, self-attention is used as a spatial attention mechanism to capture global information. Nowadays, the standard Vision Transformer usually split input images into equal-sized blocks and utilize these blocks as the token embeddings~\cite{dosovitskiy2020image}. To capture fine-grained parts' correlations, we propose the interactive channel transformation~(ICON) module in our SEMICON and utilize different channels as token embeddings. We further implement this module as a two-step computation process in order to reduce the computational complexity. 

\section{Methodology}

\subsection{Overall Framework and Notations}\label{sec:RL}
Generally, both object-level~(global-level) and part-level~(local-level) features are crucial in fine-grained visual tasks~\cite{wei2021fine}. Therefore, the overall framework of our SEMICON maintains a global feature learning branch and a local pattern learning branch, cf. Figure~\ref{fig:fw}. Correspondingly, our hash code learning component consists of two units, \ie, the global-level hash mapping unit and the local-level hash mapping unit. In particular, the global-level hash mapping unit is designed to capture object-level binary codes while the local-level hash mapping unit is additionally divided into $m$ sub linear encoder paradigms, which is beneficial to obtaining part-level binary hash codes explicitly in a stage-by-stage fashion. Thus, the final learnt hash codes contain both object-level and part-level meanings. Furthermore, our proposed suppression-enhancing mask based attention~(SEM) module and interactive channel transformation~(ICON) module are developed to generate both discriminative global-level features and correlated local-level features. 

In concretely, for each input image $\mathcal{I}$, a backbone CNN model ${\rm \Phi}_{\rm CNN}(\cdot)$ is used to extract its deep activation tensor $\bm{T}$:
\begin{equation}
\bm{T} = {\rm \Phi}_{\rm CNN}(\mathcal{I})\in \mathbb{R}^{C\times H\times W}\,.
\end{equation}
Then, based on $\bm{T}$, a global-level transforming network $\phi(\cdot)$, which is equipped with a stack of convolution layers, is performed within the global feature learning branch as:
\begin{equation}
\hat{\bm{T}} = \phi(\bm{T}; \theta_{\rm global})\in \mathbb{R}^{C'\times H'\times W'}\,,
\end{equation}
where $\theta_{\rm global}$ presents the parameters of $\phi(\cdot)$. The local pattern learning branch contains an attention guidance $\bm{P}_1 \in \mathbb{R}^{c\times H\times W}$, which is utilized to generate the attention map $\bm{M}_{1}$ in the first stage, cf. Section~\ref{sec:SEM}. With the help of the attention map, we can evaluate the attended deep descriptors in these $H\times W$ cells by conducting element-wise Hadamard product by:
\begin{equation}
\bm{T}'_{1} = \bm{M}_1 \odot \bm{T}\,.
\end{equation}
Then, the proposed SEM module is adopted to generate other attention maps $\bm{M}_{i}$ in the following $m-1$ stages, as well as the corresponding deep activation tensors $\bm{T}'_{i}$. Besides, to obtain semantic-specific representations, a local-level transforming network $\phi '(\cdot)$, which has the same structure as $\phi(\cdot)$, is used to transform $\bm{T}'_{i}$ as 
\begin{equation}
\hat{\bm{T}}'_{i} = \phi '(\bm{T}'_{i}; \theta_{\rm local})\in \mathbb{R}^{C'\times H'\times W'}\,,
\end{equation}
where $\theta_{\rm local}$ presents the parameters of $\phi '(\cdot)$. Then, the proposed ICON module is conducted over $\hat{\bm{T}}$ and $\hat{\bm{T}}'_{i}$ for making interactions across different channels.

Finally, by performing global average-pooling on $\hat{\bm{T}}$ and $\hat{\bm{T}}'_{i}$, we can obtain the object-level feature $\bm{x}^{\rm global}$ and $m$ part-level features $\bm{x}_{i}^{\rm local}$. In order to generate the binary-like codes, a binary-like code mapping module consists of $m+1$ linear encoder paradigms $\bm{W}=\{\bm{W}^{\rm global};\bm{W}^{\rm local}_1;\bm{W}^{\rm local}_2;\ldots;\bm{W}^{\rm local}_m\}$ is built to project $\bm{x}^{\rm global} / \bm{x}_{i}^{\rm local}$ as $\bm{v}^{\rm global} / \bm{v}_{i}^{\rm local}$. Eventually, the hash code learning module is performed upon $\bm{v}^{\rm global}$ and $\bm{v}_{i}^{\rm local}$ to obtain the final binary hash codes $\bm{u} = [\bm{u}^{\rm global}; \bm{u}_{1}^{\rm local}; \bm{u}_{2}^{\rm local}; \ldots ; \bm{u}_{m}^{\rm local}]$.


\subsection{Suppression-Enhancing Mask based Attention}\label{sec:SEM}
Attention in human perception renders that humans selectively focus on several salient parts of an object, which may help better capture visual structure~\cite{nipshiton2010}. Inspired by this, we incorporate the attention mechanism into the local pattern learning branch to capture the patterns of fine-grained objects' parts.

In previous fine-grained vision tasks, some works adopt the mask based attention for erasing the most discriminative regions to mine the rest of the object-specific regions in different branches~\cite{zhang2018adversarial,liu2019bidirectional}. However, the simple erasing of the most discriminative regions seems trivial and will overlook the relations between the erased regions and other significant regions. To overcome such an issue, we propose the suppression-enhancing mask based attention~(SEM) module to maintain relations among different activated regions. It is worth mentioning that the proposed SEM can be realized by convolutional layers sharing parameters, which could bring computational economy.

In concretely, for the given deep activation tensor $\bm{T}$ related to the input image $\mathcal{I}$, $m$ attention maps $\mathcal{M}=\{ \bm{M}_{1},\bm{M}_{2},\ldots,\bm{M}_{m} \}$ whose $\bm{M}_{i} \in \mathbb{R}^{H\times W}$ will be extracted. While the $m$ attention guidances ${\bm{P}}_{i}$ which are utilized to calculate the attention maps can be expressed as:
\begin{equation}
    {\bm{P}}_{i}=\left\{
    \begin{array}{c}
        \phi_{\rm att}(\bm{T};\theta_{\rm att}) {,\quad \quad \quad \ \ \ i=1}\\
        f_{SEM}(\mathrm{softmax}(\bm{M}_{i-1})) \odot {\bm{P}}_{i-1} {, \quad i=\{2,3,\ldots,m\} }
    \end{array} \right.\,,
\end{equation}
where $\phi_{att}$ is a transformation network which can be optimized in an end-to-end manner driven by the overall loss function described in Section~\ref{sec:HL} and $f_{SEM}$ is the suppression-enhancing mask based attention operation which will be described later in this section.

More specifically, the initial attention map $\bm{M}_{1}$ is generated according to the attention guidance $\bm{P}_1$ w.r.t. $\bm{T}$ in the first stage while in the following $m-1$ stages, attention maps are generated by the suppression-enhancing mask based attention operation. To obtain $\bm{M}_{1}$, a transformation network $\phi_{\rm att}$ is primarily used to obtain what to pay attention to, which can be formulated as:
\begin{equation}
{\bm{P}}_{1} = \phi_{\rm att}(\bm{T};\theta_{\rm att})\,,
\end{equation}  
where $\bm{P}_1 \in \mathbb{R}^{c\times H\times W}$ presents the attention guidance within the first stage and $\theta_{\rm att}$ presents the parameters of the corresponding network w.r.t. $\bm{T}$. Then, a $1\times 1$ convolution layer $\varphi_1$ followed by $\phi_{\rm att}$ is designed to gain $\bm{M}_1$.

For the remaining attention maps $\bm{M}_{i}$, $i=\{2,3,\ldots,m\}$ in the following $m-1$ stages, we perform the suppression-enhancing mask based attention operation $f_{SEM}$ which not only helps suppress (rather than simply erasing) the previous most discriminative region but also enhance the other activated regions. 

In details, we first calculate the weight of each cell in the attention map $\bm{M}_{i-1}$ of the previous stage by conducting a softmax function:
\begin{equation}
\bm{M}'_{{i-1}} = \mathrm{softmax}(\bm{M}_{{i-1}})\in \mathbb{R}^{H\times W}\,.
\end{equation}
Then, we record $\mu_{{i-1}}^{\rm std}$ and $\mu_{{i-1}}^{\rm mean}$ as the standard deviation value and the mean value of all the elements in $\bm{M}'_{{i-1}}$. For each element $\mu_{{i-1}}^{k}\in \{\mu_{{i-1}}^{1},\mu_{{i-1}}^{2},\ldots,\mu_{{i-1}}^{H\times W} \}$ in $\bm{M}'_{{i-1}}$, the $f_{SEM}$ operation is defined as follows:
\begin{equation}
\label{eq:alpha} \mu_{{i-1}}^{k} = 1 - \frac{\mu_{{i-1}}^{k}-\mu_{{i-1}}^{\rm mean}}{({\mu_{{i-1}}^{\rm std}})^\alpha}\,,
\end{equation}
where $\alpha$ is a hyper-parameter used to regularize the degree of suppression ratio of discriminative regions and the enhance ratio of other activated regions. Additionally, the attention guidance ${\bm{P}}_{i-1}$ of the previous stage is then changing to ${\bm{P}}_{i}$ by performing element-wise Hadamard product. The $i$th attention map is afterwards generated by the $i$th $1\times 1$ convolution layer $\varphi_{i}$. Therefore, the representations of $m$ attention maps $\bm{M}_{i}$ can be written as:
\begin{equation}
    \bm{M}_{i}=\varphi_{i}({\bm{P}}_{i}),i=\{1,2,\ldots,m\}\,.
\end{equation}
Thus, the final $m$ deep activation tensors ${\bm{T}}'_{i}$ can be obtained via
\begin{equation}
    {\bm{T}}'_{i} = \bm{M}_{i} \odot \bm{T},i=\{1,2,\ldots,m\}\,.
\end{equation}

By performing this suppression-enhancing mask based attention operation, the most discriminative region in the attention guidance of the previous stage will be partially restrained. Meanwhile, those unactivated regions will be further inhibited while other activated regions will be enhanced with attention. Therefore, relations between the activated regions of the previous stage and the activated regions generated afterwards could be maintained. 

\begin{figure}[hbt]
\centering
{\includegraphics[width=1.0\textwidth]{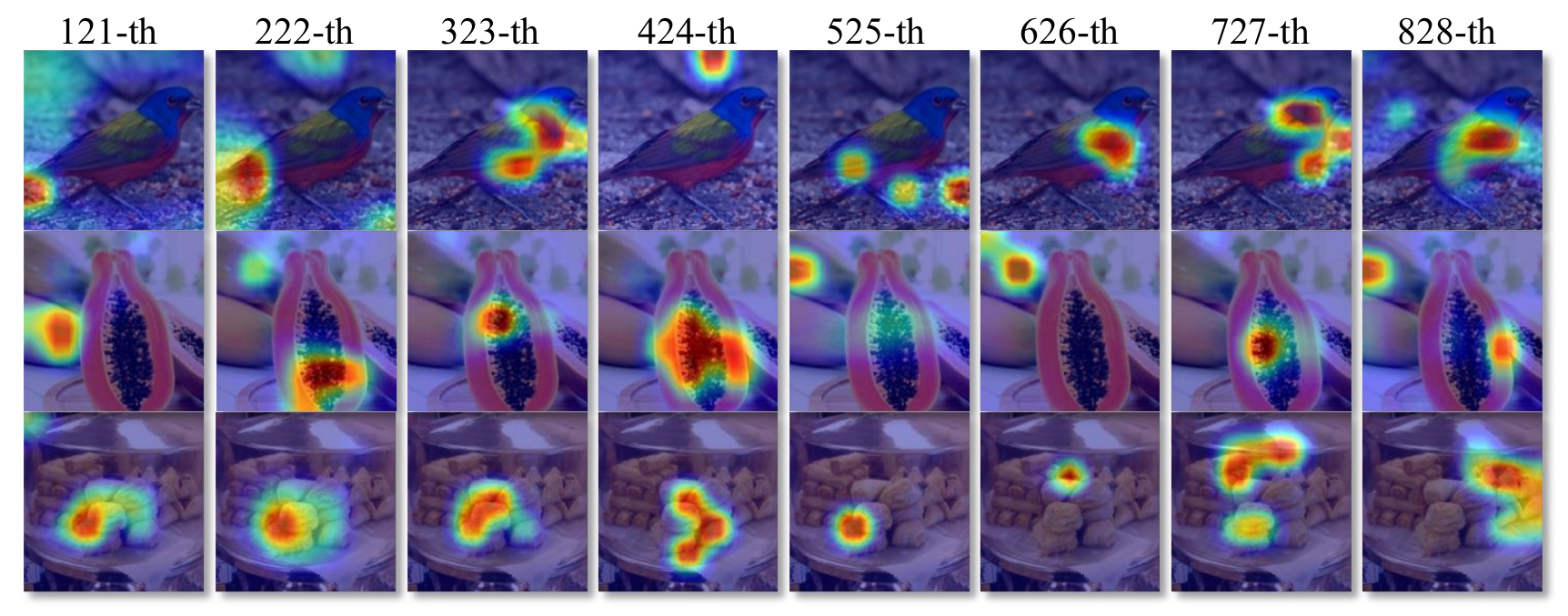}}
\caption{Visualization of channels extracted from DNNs by highlighting their weights. 
}
\label{fig:vfm}
\end{figure}

\subsection{Interactive Channel Transformation}\label{sec:ICON}
In Deep Neural Networks~(DNNs), channels are usually exploited as objects' part detectors~\cite{simon2015neural,chen2017sca,liu2015treasure}. As can be seen from Figure~\ref{fig:vfm}, the activated regions of the sampled channels~(highlighted in warm colors) are semantically meaningful. Therefore, we incorporate the self-attention mechanism into our model and utilize each channel as token embeddings to make interactions across different channels for capturing the correlations of fine-grained ``parts'', which has been proved can be greatly improved the fine-grained recognition accuracy~\cite{wei2021fine,zheng2017learning,yu2018category,chen2019destruction}.

\begin{figure}[bt]
\centering
{\includegraphics[width=1.0\textwidth]{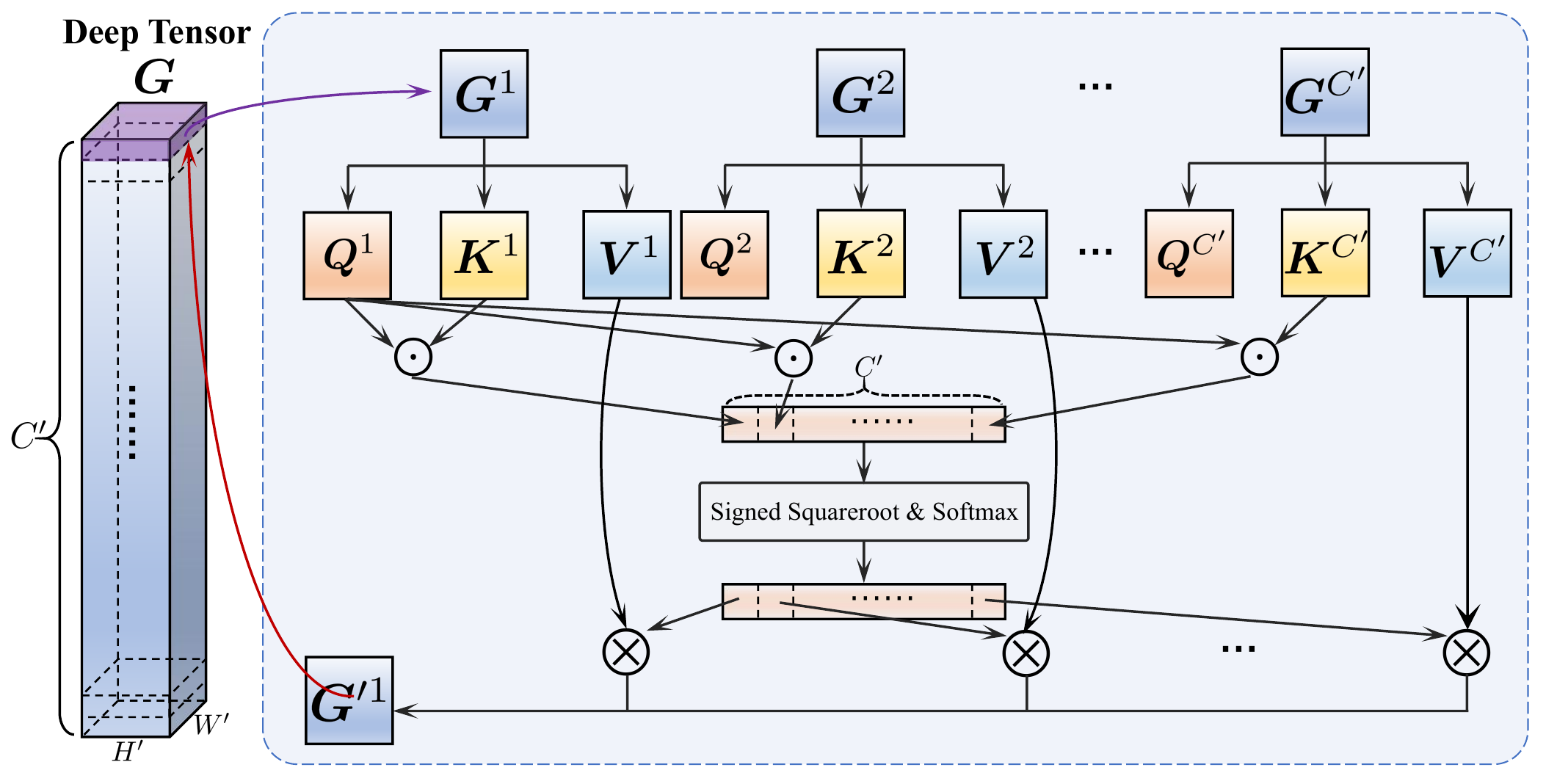}}
\caption{The \textbf{I}nteractive \textbf{C}hannel transformati\textbf{ON}~(ICON) module. It utilizes each channel as token embeddings and makes interactions across different channels.}
\label{fig:ICON}
\end{figure}

\begin{figure}[t]
\centering
{\includegraphics[width=1.0\textwidth]{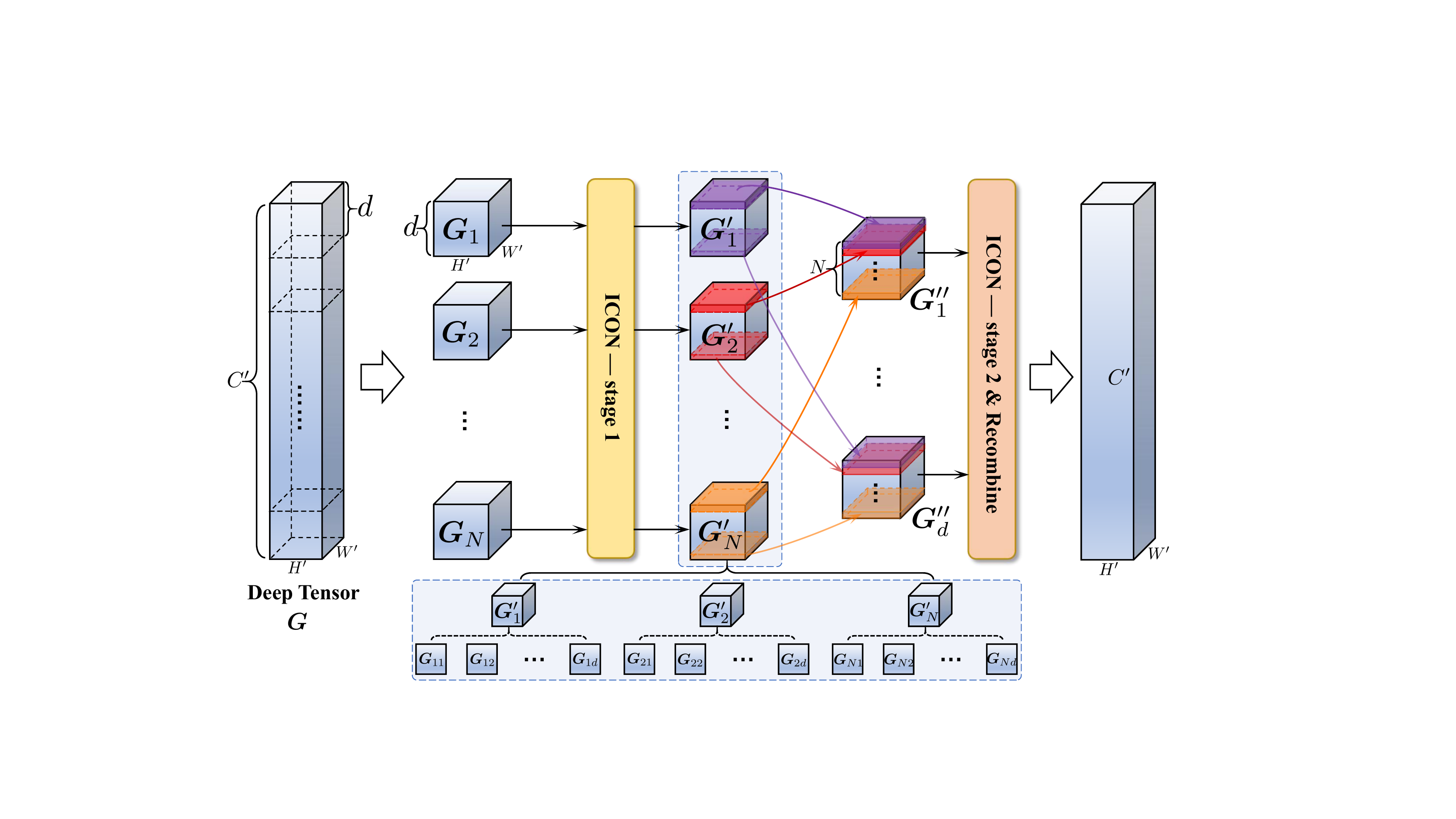}}
\caption{Our interactive channel transformation module is implemented by a two-step process for reducing the computational consumption.}
\label{fig:icon_2}
\end{figure}

In Figure~\ref{fig:ICON}, an overview of the proposed interactive channel transformation~(ICON) module is depicted. The computational complexity of directly performing the interactive channel transformation over all channels is considerable. Therefore, for the given deep tensor $\bm{G}\in \{ \hat{\bm{T}},\hat{\bm{T}}'_{1},\hat{\bm{T}}'_{2}, \ldots, \hat{\bm{T}}'_{m} \}$ of each input image $\mathcal{I}$, we split it into several portions and design a two-step interactive channel transformation module~(cf. Figure~\ref{fig:icon_2}) which can be directly adopted in traditional deep hashing frameworks to reduce the computational consumption.   

Specifically, the first step is composed of a stack of $N$ identical parts. For each given $\bm{G}$, we split the deep tensor into $N$ equal length portions $[\bm{G}_1;\bm{G}_2;\bm{G}_3;\ldots;\bm{G}_N]$, where $\bm{G}_i\in \mathbb{R}^{d\times H'\times W'}$ and $d=C'/N$. $(H',W')$ is the resolution of each channel while $C'$ is the number of channels. For each $\bm{G}_i$, the interactive channel transform operation is used to generate the transformed portion ${\bm{G}}'_i$ in order to make interactions over different channels within itself. The interactive channel transform operation during the first step can be described as mapping a unique query~($\bm{Q}_i$) and key-value~($\bm{K}_i - \bm{V}_i$) pair to an output~(${\hat{\bm{G}}}_i$), where $\bm{Q}_i$, $\bm{K}_i$, $\bm{V}_i$ are generated form $\bm{G}_i$ via a $1\times 1$ convolution layer. By following~\cite{vaswani2017attention}, we first compute the dot products $\hat{\bm{G}}_{i}$ of the query $\bm{Q}_i$ with the key $\bm{K}_i$ and divide by $\sqrt{d}$: 
\begin{equation}
\hat{\bm{G}}_{i} = \frac{\bm{Q}_i \bm{K}_i^\top}{\sqrt{d}}\,.
\end{equation} 
Then, a signed squareroot step and a softmax function is applied to generate each output ${\bm{G}}'_i$ as: 
\begin{equation}
{\bm{G}}'_i=\mathrm{softmax} \left( \mathrm{sign}(\hat{\bm{G}}_{i})\cdot \sqrt{|\hat{\bm{G}}_{i}|+\delta} \right) \bm{V}_i\,,
\end{equation} 
where $\delta$ is a fixed positive bias.  


In order to make interaction among different portions, in the second step, the tokens at the same position in $\bm{G}'_i$ are recombined into $\bm{G}''_i$. In simpler terms, for each portion ${\bm{G}}'_i=\{ {\bm{G}}_{i1};{\bm{G}}_{i2};\ldots;{\bm{G}}_{id} \}$, where
${\bm{G}}'_i \in \mathbb{R}^{d\times H'\times W'}$ and ${\bm{G}}_{ij} \in \mathbb{R}^{H'\times W'}$ obtained from the first step, we recombine these portions by integrating those channels with the same index in preparation for the second step interactive channel transformation. To be specific, the recombined portion ${\bm{G}}''_i$ is consisted of $N$ channels from the previous $N$ portions ${\bm{G}}'_i$:
\begin{equation}
    {\bm{G}}''_i = \{ {\bm{G}}_{1i};{\bm{G}}_{2i};\ldots;{\bm{G}}_{Ni} \}, i=\{1,2,\ldots,d\}\,.
\end{equation}
The second step ICON is then performed on ${\bm{G}}''_i$ with the same processes as the first step. Finally, channels which have changed their original index will be reset after performing the two-step ICON process.

Between these two steps, we employ a batch normalization and a ReLU activation. A residual connection~\cite{resnet16} is adopted after each step. Instead of performing a single interactive channel transform operation associated with keys, values and queries, inspired by~\cite{vaswani2017attention}, we perform the two-step interactive channel transform operation in parallel. For traditional deep hashing frameworks generally use CNNs as vanilla backbones, we perform group convolutions as a substitute for multi-head linear projections. The group number across the first step is $N$ while it will be reset as $d$ within the second step. This two-step multi-group interactive channel transformation allows the model to jointly process information within different indexes over different channels.

\subsection{Hash Code Learning}\label{sec:HL}
In the following, we conduct the hash code learning based on the obtained \mbox{object-level} features and part-level features. Assuming that we have $q$ query data points which are denoted as $\{\bm{q}_i\}_{i=1}^{q}$, as well as $p$ database points which are donated as $\{\bm{p}_j\}_{j=1}^{p}$. For each $\bm{q}_i$ and $\bm{p}_j$, it consists of a global feature $\bm{v}^{\rm global}$ and $m$ local features $\bm{v}^{\rm local}_i$. The corresponding hash codes can be carried out by
\begin{eqnarray}
\bm{u}_i = \mathrm{sign}({\bm{q}_i})\,,\quad \bm{z}_j = \mathrm{sign}({\bm{p}_j})\,, 
\end{eqnarray}
where $\bm{u}_i,\bm{z}_j\in \{-1,+1\}^k$, and $k$ presents the length of the final binary hash codes. The goal of hashing is to learn binary hash codes for both query points and database points and preserving their similarity simultaneously. Following~\cite{qingyuanAAAI18}, the formulation of the hash code learning can be written as:
\begin{align}
\label{eq:squaredhash} \min_{\bm{W},\Theta} \mathcal{L}(\mathcal{I}) = \sum_{i\in\Omega} \sum_{j\in\Gamma} \left[ {\mathrm{sign}(\bm{W}\cdot F(\mathcal{I}_i;\Theta))}^\top \bm{z}_j - k S_{ij}\right]^2,\quad \bm{z}_j\in\{-1,+1\}^k,
\end{align}
where $\Gamma$ presents the indices of all the database points while $\Omega \subseteq \Gamma$ presents the indices of the query set points for we can only gain access to a set of database points $\{\bm{p}_j\}_{j=1}^p$ without query points during the training stage. $S\in \{-1,+1\}^{q\times p}$ denotes the pairwise supervised information. $\bm{W}$ presents the matrix of $m+1$ linear projection and $\Theta$ denotes the parameters of DNNs to be learned.

By relaxation, we get the final formulation of SEMICON:
\begin{align}
\label{eq:final} \min_{\bm{W},\Theta} \mathcal{L}(\mathcal{I}) = &\beta \sum_{i\in\Omega} \sum_{j\in\Gamma} \left[ {\tanh(\bm{W}\cdot F(\mathcal{I}_i;\Theta))}^\top \bm{z}_j - k S_{ij}\right]^2 \nonumber \\
&+\gamma \sum_{i\in\Omega}\left[\bm{z}_i - \tanh(\bm{W}\cdot F(\mathcal{I}_i;\Theta)) \right]^2  \,,
\end{align}
where $\beta$ and $\gamma$ are hyper-parameters as the trade-off. The proposed SEMICON is an end-to-end deep hashing method which is able to simultaneously perform feature learning and hash code learning in such a unified framework.

\section{Experiments}

\subsection{Datasets}
By following \textsc{A$^2$-Net}~\cite{wei20212} and ExchNet~\cite{exchnet}, our experiments are conducted on two widely used fine-grained datasets, \ie, \textit{CUB200-2011}~\cite{WahCUB200_2011} and \textit{Aircraft}~\cite{airplanes}, as well as three popular large-scale fine-grained datasets, \ie, \textit{Food101}~\cite{food101}, \textit{NABirds}~\cite{nabirds15} and \textit{VegFru}~\cite{vegfru}. Specifically, \textit{CUB200-2011} contains 11,788 bird images from 200 bird species and is officially split into 5,994 images for training and 5,794 images for test. \textit{Aircraft} contains 10,000 images of 100 aircraft variants, among which 6,667 images for training and 3,333 images for test. For large-scale datasets, \textit{Food101} contains 101 kinds of foods with 101,000 images, where for each class, 250 test images are checked manually for correctness while 750 training images still contain a certain amount of noises. \textit{NABirds} contains 48,562 images of North American birds with 555 sub-categories, 23,929 images for training while 24,633 images for test. \textit{VegFru} is another large-scale fine-grained dataset covering 200 kinds of vegetables and 92 kinds of fruits with 29,200 for training, 14,600 for validation and 116,931 for test.

\subsection{Baselines and Implementation Details}\label{sec:BID}

\paragraph{\rm \textbf{Baselines}}
In experiments, we compare our proposed method to the following competitive generic hashing methods, \ie, ITQ~\cite{gong2012iterative}, SDH~\cite{shen2015supervised}, DPSH~\cite{li2015feature}, HashNet~\cite{cao2017hashnet}, and ADSH~\cite{qingyuanAAAI18}. Among them, DPSH, HashNet and ADSH are also deep learning based methods, while ITQ and SDH are not. Furthermore, we also compare the results of our SEMICON with state-of-the-arts of fine-grained hashing methods, including ExchNet~\cite{exchnet} and \textsc{A$^2$-Net}~\cite{wei20212}. 

\paragraph{\rm \textbf{Implementation Details}}
For fair comparisons, we follow the training setting in \textsc{A$^2$-Net}~\cite{wei20212} and ExchNet~\cite{exchnet}. In concretely, for \textit{CUB200-2011}, \textit{Aircraft} and \textit{Food101}, we only sample 2,000 images per epoch for training, while 4,000 samples are randomly selected per epoch for \textit{NABirds} and \textit{VegFru}. For the training details, regarding the backbone model, we can choose any network structures as the base network for fine-grained representation learning. While, by following ExchNet~\cite{exchnet} and \textsc{A$^2$-Net}~\cite{wei20212}, ResNet-50~\cite{resnet16} is employed in experiments for fair comparisons. The attention generation network $\phi_{\rm att}$ is the fourth stage of ResNet-50 without downsample convolutions. The global-level transforming network $\phi(\cdot)$ and the local-level transforming network $\phi '(\cdot)$ are independent networks, sharing the same architecture with the fourth stage of ResNet-50. The total number of training epochs is $30$. The iteration time is $40$ for those datasets containing less than 20,000 training images while for other datasets, the iteration time is $50$. For all datasets, we preprocess all images to $224\times 224$, and the learning rate is set to $2.5\times 10^{-4}$ for all iterations. SGD with mini-batch set as $16$ is used for training. We set the weight decay as $10^{-4}$ and momentum as $0.91$. The hyper-parameters, \ie, $\alpha$ in Eq.~\eqref{eq:alpha} and $\beta$, $\gamma$ in Eq.~\eqref{eq:final}, are set as $0.3$, $1$ and $200$, respectively. By following ADSH~\cite{qingyuanAAAI18}, we adopt soft-constraints strategy~\cite{leng2014supervised} to avoid the similarity imbalance problem. The number of $m$ is set as $3$ which means there exists $3$ attention maps $\bm{M}_i$. The length of the final hash code $\bm{u}^{\rm global}$ and $\bm{u}^{\rm local}_{i}$ is set as $\lceil \frac{k}{2} \rceil$ and $\lfloor \frac{k}{6} \rfloor$. The fixed positive bias $\delta$ is set as $10^{-5}$. All experiments are conducted with one GeForce RTX 2080 Ti GPU.
\begin{table}[t]
\renewcommand{\arraystretch}{1.1}
\centering
\caption{Comparisons of retrieval accuracy (\% mAP) on five fine-grained datasets.}
\begin{tabular}{c|c|rrrrrrr|r}
\toprule
Datasets & \multicolumn{1}{c|}{\# bits} & \multicolumn{1}{c}{ITQ} & \multicolumn{1}{c}{SDH} & \multicolumn{1}{c}{DPSH} & \multicolumn{1}{c}{HashNet} & \multicolumn{1}{c}{ADSH} & \multicolumn{1}{c}{ExchNet} & \multicolumn{1}{c|}{\textsc{A$^2$-Net}} &\multicolumn{1}{c}{\textbf{Ours}} \\
\hline
\multirow{4}[2]{*}{\textit{CUB200-2011}} & 12 & 6.80 & 10.52 & 8.68  & 12.03 & 20.03 & 25.14 & 33.83 & \textbf{37.76}\bigstrut[t]\\
& 24 & 9.42 & 16.95 & 12.51 & 17.77 & 50.33 & 58.98 & 61.01 & \textbf{65.41}\\
& 32 & 11.19 & 20.43 & 12.74 & 19.93 & 61.68 & 67.74 & 71.61 & \textbf{72.61}\\
& 48 & 12.45 & 22.23 & 15.58 & 22.13 & 65.43 & 71.05 & 77.33 & \textbf{79.67}\\
\hline
\multirow{4}[2]{*}{\textit{Aircraft}} & 12 & 4.38 & 4.89 & 8.74 & 14.91 & 15.54 & 33.27 & 42.72 & \textbf{49.87}\bigstrut[t]\\
& 24 & 5.28 & 6.36 & 10.87 & 17.75 & 23.09 & 45.83 & 63.66 & \textbf{75.08}\\
& 32 & 5.82 & 6.90 & 13.54 & 19.42 & 30.37 & 51.83 & 72.51 & \textbf{80.45}\\
& 48 & 6.05 & 7.65 & 13.94 & 20.32 & 50.65 & 59.05 & 81.37 & \textbf{84.23}\\
\hline
\multirow{4}[2]{*}{\textit{Food101}} & 12 & 6.46 & 10.21 & 11.82 & 24.42 & 35.64 & 45.63 & 46.44 & \textbf{50.00}\bigstrut[t]\\
& 24 & 8.20 & 11.44 & 13.05 & 34.48 & 40.93 & 55.48 & 66.87 & \textbf{76.57} \\
& 32 & 9.70 & 13.36 & 16.41 & 35.90 & 42.89 & 56.39 & 74.27 & \textbf{80.19} \\
& 48 & 10.07 & 15.55 & 20.06 & 39.65 & 48.81 & 64.19 & 82.13 & \textbf{82.44}\\
\hline
\multirow{4}[2]{*}{\textit{NABirds}} & 12 & 2.53 & 3.10 & 2.17 & 2.34 & 2.53 & 5.22 & \textbf{8.20} & 8.12\bigstrut[t]\\
& 24 & 4.22 & 6.72 & 4.08 & 3.29 & 8.23 & 15.69 & 19.15 & \textbf{19.44}\\
& 32 & 5.38 & 8.86 & 3.61 & 4.52 & 14.71 & 21.94 & 24.41 & \textbf{28.26} \\
& 48 & 6.10 & 10.38 & 3.20 & 4.97 & 25.34 & 34.81 & 35.64 & \textbf{41.15} \\
\hline
\multirow{4}[2]{*}{\textit{VegFru}} & 12 & 3.05 & 5.92 & 6.33 & 3.70 & 8.24 & 23.55 & 25.52 & \textbf{30.32}\bigstrut[t]\\
& 24 & 5.51 & 11.55 & 9.05 & 6.24 & 24.90 & 35.93 & 44.73 & \textbf{58.45}\\
& 32 & 7.48 & 14.55 & 10.28 & 7.83 & 36.53 & 48.27 & 52.75 & \textbf{69.92} \\
& 48 & 8.74 & 16.45 & 9.11 & 10.29 & 55.15 & 69.30 & 69.77 & \textbf{79.77} \\	
\bottomrule
\end{tabular}%
\label{table:results1}%
\end{table}%

\subsection{Main Results}
Table~\ref{table:results1} presents the mean average precision (mAP) results of fine-grained retrieval for comparisons with state-of-the-art hashing methods on these five aforementioned benchmark fine-grained datasets. For each dataset, we report the results of four lengths of hash bits, \ie, 12, 24, 32, and 48, for evaluations. From table~\ref{table:results1}, we can observe that the proposed SEMICON significantly outperforms the other baseline methods on these datasets. 

In particular, compared with the state-of-the-art method \textsc{A$^2$-Net}~\cite{wei20212}, our SEMICON achieves 11.42\% and 17.17\% improvements over \textsc{A$^2$-Net} of 24-bit and 32-bit experiments on \emph{Aircraft} and \emph{VegFru}. Moreover, SEMICON obtains superior results on both medium-scale fine-grained datasets, \eg, \emph{CUB200-2011} and \emph{Aircraft}, and large-scale fine-grained datasets, \eg, \emph{NABirds} and \emph{VegFru}. These observations validate the effectiveness of the proposed SEMICON, as well as its promising practicality in real applications of fine-grained retrieval. 

\begin{table}[htbp]
  \centering
  \begin{threeparttable}
  \scriptsize
  \renewcommand{\arraystretch}{1.2}
  \caption{\scriptsize Retrieval accuracy (\% mAP) with incremental modules of the proposed SEMICON.}
    \begin{tabular}{c|cccc|cccc|cccc}
    \toprule
    \multirow{2}[1]{*}{Configurations} & \multicolumn{4}{c|}{\textit{CUB200-2011}}      & \multicolumn{4}{c|}{\textit{Aircraft}} & \multicolumn{4}{c}{\textit{Food101}} \\
          & 12  & 24  & 32  & 48  & 12  & 24  & 32 & 48 & 12 & 24 & 32 & 48 \\
    \hline
    Vanilla backbone & 20.03  & 50.33  & 61.68  & 65.43  & 15.54  & 23.09  & 30.37  & 50.65  & 35.64  & 40.93  & 42.89  & 48.81  \bigstrut[t]\\
    + SEMICON$^{-*}$ & 34.93  & 58.73  & 64.71  & 75.66  & 34.18  & 70.14  & 76.50  & 80.23  & 40.59  & 72.75  & 78.98  & 80.15  \\
    + SEM  & 36.58  & 64.19  & 71.58  & 79.17  & 43.36  & 73.39  & \textbf{80.64}  & 83.99  & 44.95  & 75.44  & 80.07  & 82.40  \\
    + ICON & \textbf{37.76}  & \textbf{65.41}  & \textbf{72.33}  & \textbf{79.62}  & \textbf{49.87}  & \textbf{75.08}  & 80.45  & \textbf{84.23}  & \textbf{50.00}  & \textbf{76.57}  & \textbf{80.19}  & \textbf{82.44}  \\
    \bottomrule
    \end{tabular}
    \label{table:results2}%
    \fontsize{7pt}{2.4mm}
    \begin{tablenotes}
        \item[*] SEMICON$^{-}$ represents the model generates $m$ attention maps without performing SEM and the proposed ICON is not performed before obtaining the final hash codes.
      \end{tablenotes}
  \end{threeparttable}
\end{table}%

\subsection{Ablation Studies}

We demonstrate the effectiveness of these crucial modules of the proposed SEMICON, \ie, the novel hash learning framework (cf. Section~\ref{sec:RL}), the suppression-enhancing mask based attention~(SEM) module (cf. Section~\ref{sec:SEM}) and the interactive channel transformation~(ICON) module (cf. Section~\ref{sec:ICON}). In the ablation studies, we apply these modules incrementally on a vanilla backbone (\ie, ResNet-50) as the baseline. As evaluated in Table~\ref{table:results2}, by stacking these modules one by one, the retrieval results are steadily improved, which justifies the effectiveness of our proposals in SEMICON.

\section{Conclusion}\label{sec:conc}
In this paper, we proposed the Suppression-Enhancing
Mask based Attention and Interactive Channel Transformation (SEMICON) for dealing with the large-scale fine-grained image retrieval task. In concretely, the SEM module was developed to restrain (rather than simply erasing) the most discriminative region under the attention guidance of the previous stage, which benefited maintaining relations between different activated regions in a stage-by-stage fashion. Moreover, as channels in DNNs could often correspond to object parts, our ICON module treated each channel as token embeddings for capturing fine-grained parts' correlations. With the hash mapping component containing two units of both global-level and local-level, the final learnt binary hash codes can be generated from different features with different levels~(\ie, global-level and local-level) respectively. Experiments on five fine-grained datasets demonstrated the effectiveness of our SEMICON, as well as its proposals. In the future, we would like to improve the robustness of hashing methods and conduct experiments under a more generalized retrieval setting where training classes and test classes have no overlap.

\section*{Acknowledgements}
The authors would like to thank the anonymous reviewers for their critical and constructive comments and suggestions. We gratefully acknowledge the support of MindSpore, CANN (Compute Architecture for Neural Networks) and Ascend AI Processor used for this research.

{
    \bibliographystyle{splncs04}
    \bibliography{egbib}
}

\end{document}